# Auditing ICU Readmission Rates in an Clinical Database: An Analysis of Risk Factors and Clinical Outcomes

Shaina Raza
*Vector Institute of Artificial Intelligence*
Toronto, Canada
shaina.raza@vectorinstitute.ai

*Abstract*— **This study presents a machine learning (ML) pipeline for clinical data classification in the context of a 30-day readmission problem, along with a fairness audit on subgroups based on sensitive attributes. A range of ML models are used for classification and the fairness audit is conducted on the model predictions. The fairness audit uncovers disparities in equal opportunity, predictive parity, false positive rate parity, and false negative rate parity criteria on the MIMIC III dataset based on attributes such as gender, ethnicity, language, and insurance group. The results identify disparities in the model's performance across different groups and highlights the need for better fairness and bias mitigation strategies. The study suggests the need for collaborative efforts among researchers, policymakers, and practitioners to address bias and fairness in artificial intelligence (AI) systems.**

*Keywords—Fairness, bias, health, disparity, clinical data.*

## I. INTRODUCTION

The intensive caring unit (ICU) is an essential unit for the management of critically ill patients. However, many of these patients are at risk of readmission within 30 days of discharge, which can result in additional burdens on the patient, their family, and the healthcare system [1], [2]. As a result, the development of a model that accurately predicts readmission risk can assist clinicians in identifying high-risk patients and intervening early to prevent readmission. However, there is a possibility of bias in such models due to factors such as race, gender, and socioeconomic status, which can lead to healthcare disparities and exacerbate existing health inequities [3]. Research [4], [5] has shown that some predictive models may be less accurate in predicting health outcomes for certain racial and ethnic groups, exacerbating existing health disparities. In this context, bias [6], [7] refers to the presence of systematic errors in the data or model that lead to unfair, inaccurate, or unequal treatment of certain groups of patients.

To explore these issues, we have considered the hospital readmission use case on the MIMIC III dataset [8]. This dataset contains comprehensive records of over 60,000 ICU admissions for approximately 40,000 patients over a ten-year period, making it an useful resource for developing and testing predictive models for readmission [2], [9]. Recent studies [3], [10], [11] have discussed that the datasets used for predictive healthcare models often contain biases related to factors such as race, gender, and insurance status. These biases have the potential to negatively impact the accuracy of machine learning (ML) models and contribute to healthcare disparities. In this study, we want to explore the existence of systematic biases in a clinical dataset that may impact the readmission outcomes among the patients.

Our study aims to identify and quantify the biases, particularly those that are replicated by ML models. By doing so, we can develop strategies to address these biases and create more equitable predictive models for ICU readmission. The findings of this paper can be valuable to clinicians and researchers working to develop predictive models, as well as those working to reduce healthcare disparities and promote health equity.

## II. RELATED WORK

In recent years, the application of ML and AI in the healthcare industry has garnered significant interest. While these technologies offer many benefits, such as fast processing, automated decision-making, there is also a growing concern that they may reinforce existing biases in healthcare, perpetuating health disparities [7]. For instance, a widely used algorithm for managing high-risk patients was found to be less accurate for Black patients than their White counterparts in one study [12]. Similarly, another study [13] found that an algorithm for allocating additional care resources to patients with chronic kidney disease disproportionately excluded certain demographic groups.

Several studies have explored the issue of fairness and equity in ML application in healthcare. Some have emphasized the need for inclusive research guidelines and technical solutions to tackle biases in ML models [14]–[16]. Others have discussed the potential implications of unmitigated bias in AI-driven healthcare [4], [6], [17], highlighting discriminatory practices in the use of big data in public health research. There is also growing literature [18], [19] on specific areas of healthcare, such as ML algorithms in the health insurance industry, policy-making, and resource allocation decisions.

Despite so much technical considerations, there is still a lack of engagement with issues of fairness, bias, and interpretability in ML research in healthcare. This paper specifically focuses on interpreting biases in the data that are exhibited in model predictions and highlights the need for greater attention to issues of fairness and equity in the use of ML in healthcare.

## III. METHODOLOGY

### A. Preliminaries

The key terms used in this study is taken from the pertinent literature [20]–[22] and given below:
- *Bias*: A systematic error in the collection, interpretation, or presentation of data that can lead to incorrect conclusions.
- *Fairness:* The absence of discrimination, favoritism, or bias in the treatment of individuals or groups.
- *Public health equity*: Ensuring that everyone has a fair and just opportunity to be as healthy as possible, regardless of their race, ethnicity, socioeconomic status, or other factors.
- *Sensitive/protected attribute*: A characteristic of an individual, such as race, gender, or age, that should not be used as a basis for discrimination or differential treatment.
- *Privileged group*: A group of individuals that has systematic advantage in society based on their race, gender, socioeconomic status, or other factors.
- *Underprivileged group*: A group of individuals that has systematic disadvantage in society based on their race, gender, socioeconomic status, or other factors.

### B. Problem Definition

The problem defined in this study is two-fold: first, predicting the likelihood of intensive care readmission within 30 days of discharge, using patient release records; and second, ensuring that the model is thoroughly audited to identify and mitigate any potential sources of bias or discrimination. The audit process involves evaluating the model's performance on various patient subgroups, such as those based on race, gender, and other relevant criteria. This helps to detect any instances of bias that may unfairly favor or disadvantage certain subgroups. While the first part of problem definition (prediction 30-days readmission) [1], [2] is a common task in medical informatics, the second represents a novel and unique contribution of this study. We show the problem to solve in the large context of a ML pipeline in Figure 1.

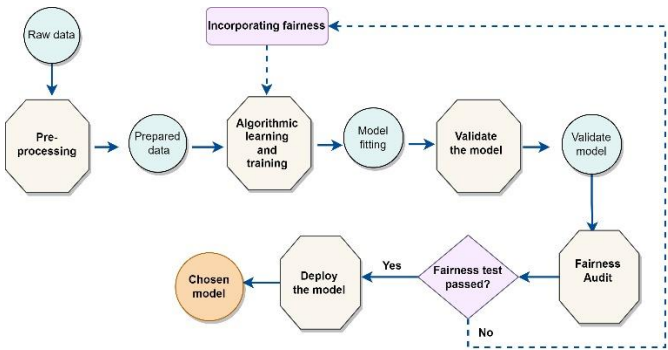

Fig. 1. Proposed method in the context of pipeline approach

As shown in Figure 1, the ML pipeline consists of several steps: data preprocessing, model training and validation, and fairness audit (new process). During the fairness audit, the model's predictions are examined for any discriminatory biases based on sensitive attributes. If biases are detected, a fairness algorithm can be applied to debias the data, and the model can be retrained and re-evaluated before deployment. The focus of this work is on the fairness audit, which involves examining the model's predictions for discriminatory biases. While a fairness algorithm is mentioned as a possible solution if biases are detected, it is not within the scope of this paper.

### C. Models

We address the 30-day intensive care readmission problem by training a range of classifiers, including Naïve Bayes (NB), Generalized Linear Model (GLM), Logistic Regression (RF), and Multi-Layer Perceptron (MLP), on the dataset and features described in the next section. The hyperparameters used to train these models are also given in Section IV (B).

### D. Fairness Audit

A fairness audit [23], [24] is a critical process that involves a series of steps to assess the fairness of a ML model. The first step in conducting a fairness audit is to identify the sensitive attributes in the data that may affect fairness, such as race, gender, or age. Analyzing the distribution of these attributes across the data can help identify potential imbalances that may lead to bias. The next step is to evaluate the performance of the model across different groups by calculating metrics such as true positive rates, false positive rates, and accuracy for each group. This step helps identify any disparities in performance that may be caused by inherent biases in the model or data. Once potential biases are identified, the next step is to use some techniques to mitigate them. This may involve adjusting the model to account for imbalanced data or introducing regularization techniques to prevent overfitting, methods such as data augmentation or re-sampling can also be used to improve the representation of underrepresented groups in the outcomes.

## IV. EXPERIMENTAL SETUP

### A. Data

The MIMIC III [25] dataset contains a wide range of demographic information on patients. Our initial analysis of the dataset reveals some of the demographic statistics, which are: the age range of patients spans from newborns to elderly individuals, with a median age of 65 years. About a quarter of the patients are under 50 years old. In terms of gender, approximately 55% of patients are male and 45% are female. The majority of patients in the dataset are White (around 69%), with representation from other races and ethnicities such as Black/African American, Hispanic/Latino, Asian, and other/mixed races. Information on socioeconomic status is not explicitly provided, but it can be inferred from variables such as insurance type (Medicare, private insurance, self-pay, government insurance, other/unknown) and zip code.

The study population for this research consists of patients who are over 18 years old and have had their first ICU stay as the index hospitalization, in cases where a patient has had multiple hospital stays. Patients who died during their hospitalization or were transferred between hospital units are excluded from the study. The exclusion criteria were applied to ensure that the study population consisted of patients who were at risk of readmission and were discharged alive from the ICU without being transferred to another unit, which could impact their readmission risk. The final dataset, after all these filtration criteria, consisted of 6.5K examples split evenly between positive examples and negative examples.

We have identified a set of features that have been shown to be useful for a 30-days readmission model in recent works [26]–[28]. These features are:

- *Demographic variables:* age, gender, race, and ethnicity.
- *Length of stay*: The number of days the patient stayed in the hospital during the index admission.
- *Admission source*: The source of the patient's admission to the hospital, such as emergency department, transfer from another facility, or elective admission.
- *Primary diagnosis:* The main reason for the patient's admission to the hospital, identified using ICD-9 codes.
- *Comorbidities:* The presence of any chronic or acute conditions other than the primary diagnosis, identified using ICD-9 codes.
- *Laboratory values*: Various laboratory values, including blood glucose levels, serum creatinine levels, and white blood cell count, were included as features.
- *Vital signs:* Vital signs, such as heart rate, respiratory rate, and blood pressure, were included as features.
- *Medication use:* Information on the use of medications during the index hospitalization was included as a feature.

The ground truth label for 30-day ICU readmission is not provided by the dataset but we calculate it using the binary readmission label linked to each hospital admission. From a technical perspective, we used a combination of demographic variables, length of stay, admission source, primary diagnosis, comorbidities, laboratory values, vital signs, and medication use as features to train a ML model for predicting the readmissions.

### B. Hyperparameters

To train the models, we explored different hyperparameters for each classifier. For NB, we varied the smoothing parameter to avoid probabilities of zero. For LG, we used L2 regularization. For the GLM, we varied the distribution and link functions. For MLP, we experimented with 2 and 3-layer models, varying the number of neurons per layer from 8 to 128 and using ReLU and tanh activation functions. We also tried different learning rate schedulers (constant and adaptive) and optimizers (Adam and Stochastic Gradient Descent) and report the best results for each model. The dataset is divided into a 70/30 train-test split. The model was trained using hyperparameter optimization, with 5-fold cross-validation used to find the best model parameters. During each fold of the cross-validation process, 80% of the training data was used for training and 20% was used for validation.

### C. Evaluation Metrics

To compare our model's performance for the classification task, we use precision, recall and F1-score as the evaluation metrics. To ensure fairness in ML models, various parity metrics are utilized based on the terms of the confusion matrix, including True Positives (TP), False Positives (FP), False Negatives (FN), and True Negatives (TN). These definitions of these fairness evaluation measures are calculated based on the literature and tools from [24], [29] and are given in Table I:

TABLE I: FAIRNESS METRICS FOR EVALUATING MACHINE LEARNING MODELS BASED ON CONFUSION MATRIX TERMS.

| Parity Metric | Definition |
|---|---|
| Equal Parity (EP) [30] | Ensures equal representation of sensitive groups. The Predicted Positive Rate (PPR) disparity for each group should be within the fairness threshold selected (.8 to 1.25). PPR can be defined as: (TP + FP) / (TP + FP + TN + FN). |
| Proportional Parity (PP) [29] | Ensures proportional representation. It measures the disparity between the Predicted Positive Group Rates (PPGR) of each group. PPGR can be defined as (TP+FP)/(TP+FP+TN+FN) for each group. |
| False Positive Rate Parity (FPRP) [29] | Ensures the same false positive rate as the reference group. It measures the disparity between the False Positive Rates (FPR) of each group. FPR can be defined as FP/(FP+TN). |
| False Negative Rate Parity (FNRP) [29] | Ensures the same false negative rate as the reference group. It measures the disparity between the False Negative Rates (FNR) of each group. FNR can be defined as FN/(FN+TP). |

In this study, we have defined sensitive groups based on insurance type (government, private, Medicare, Medicaid, and self-pay), gender (male, female), ethnicity (White, Black, Hispanic, Asian, and others), and language (English as primary language and non-English speakers). To simplify our analysis, we grouped individuals by main categories for ethnicity and language (there were numerous subgroups within each category). A reference group [24] is used as the baseline to evaluate bias during the fairness evaluation, with 'male,' 'white,' 'Medicare,' and 'English speaker' as the reference groups for gender, ethnicity, insurance, and language respectively. These groups are chosen as they are the majority group with most favorable outcomes (i.e., 30-day readmission). Following the industry standard for fairness [31]–[33], if disparity for a group is within 80% and 125% of the value of the reference group on a group metric (e.g. FPR or so), this audit test is considered to pass.

## V. RESULTS AND ANALYSIS

### A. Classification Results

Table II shows the models' results for a 30-day readmission problem on the MIMIC III dataset.

TABLE II: CLASSIFICATION TASK PERFORMANCE ON MIMIC DATASET.

| Model | Precision | Recall | F1 Score |
|---|---|---|---|
| NB | 0.80 | 0.75 | 0.77 |
| GLM | 0.82 | 0.76 | 0.79 |
| LR | 0.85 | 0.78 | 0.81 |
| MLP | **0.87** | **0.81** | **0.84** |

Overall, the results in Table II shows that all four models perform relatively well in terms of precision, recall, and F1 score, with MLP showing the best overall performance. LR and GLM also perform well, while NB has a slightly lower F1 score. Nevertheless, the main goal of this study is to evaluate the classification results for fairness metrics. Due to its best performance, we choose the results of MLP for the bias and fairness assessment task.

*B. Assessing Bias*

Results for assessing bias and fairness on model (MLP) classification are given in Table III and discussed next.

TABLE III: FAIRNESS EVALUATION USING PREDICTED POSITIVE RATE (PPR), PREDICTED POSITIVE GROUP RATE (PPGR), FALSE POSITIVE RATE (FPR) AND FALSE NEGATIVE RATE (FNR). THESE VALUES ARE USED TO CALCULATE THE EQUAL PARITY (EP), PROPORTIONAL PARITY (PP), FALSE POSITIVE RTE PARITY (FPRP), FALSE NEGAIVE RATE PARITY (FNRP).

| Group (size ratio) | PPR | PPGR | FPR | FNR |
|---|---|---|---|---|
| Insurance | | | | |
| Government (0.04) | 0.01 | 0.09 | 0.07 | 0 |
| Medicaid (0.12) | 0.09 | 0.23 | 0.19 | 0.06 |
| Medicare (0.43) | 0.55 | 0.40 | 0.37 | 0.08 |
| Private (0.39) | 0.34 | 0.27 | 0.25 | 0.14 |
| Self pay (0.02) | 0 | 0.04 | 0.02 | 0 |
| Gender | | | | |
| F (0.43) | 0.35 | 0.25 | 0.22 | 0.12 |
| M (0.57) | 0.65 | 0.35 | 0.32 | 0.07 |
| Ethnicity | | | | |
| Asian (0.03) | 0.01 | 0.08 | 0.04 | 0 |
| Black (0.11) | 0.12 | 0.32 | 0.27 | 0.02 |
| Hispanic (0.05) | 0.02 | 0.11 | 0.07 | 0.02 |
| Other (0.13) | 0.02 | 0.04 | 0.02 | 0.05 |
| White (0.69) | 0.85 | 0.37 | 0.35 | 0.12 |
| Language | | | | |
| English (0.67) | 0.8 | 0.37 | 0.34 | 0.1 |
| Non-English (0.33) | 0.2 | 0.18 | 0.16 | 0.07 |

*Impact on Equal Parity (EP):* The results presented in Table III indicate that the classification model did not satisfy the EP criterion. This means that some groups were not equally represented in the selected set. Specifically, the EP disparity for the Medicaid group is calculated to be 0.16X lower than the reference group (Medicare), the Government group is 0.02X lower, the Self-Pay group is 0.001X lower, and the Private group is 0.63X higher than the reference group (X represents the degree of disparity between the reference and other groups). These disparities demonstrate that certain groups were underrepresented (Medicaid, government, self-pay) or overrepresented (Private) in the selected set, indicating a lack of equal treatment in the classification model.

For the attribute Gender, with the reference group being Male, the EP disparity for the Female group is found to be 0.53X. This means that the proportion of females in the selected set is 0.53 times the proportion of males in the same set, indicating that females are underrepresented in the selected set.

For the Ethnicity Group, with the reference group being White, the EP disparities for the Other, Black/African American, Hispanic, and Asian groups are found to be 0.02X, 0.14X, 0.02X, and 0.01X, respectively. These disparities indicate that certain ethnic groups were underrepresented in the selected set, with the Black group having the largest disparity of 0.14X.

For the attribute Language Group, with the reference group being English Speaker, the EP disparity for the Non-English group was found to be 0.25X, indicating that the Non-English group was underrepresented in the selected set compared to the English Speaker group.

*Impact on Proportional Parity (PP):* The PP criteria measures whether each group is represented proportionally to its share of the population. In Table III, the result shows that several groups failed the PP criteria. For Insurance group, the Self Pay group is underrepresented with a disparity of 0.10X, while the Private insurance group is overrepresented with a disparity of 0.67X. For Gender, the Female group is underrepresented with a disparity of 0.70X. For Ethnicity, the Other group is underrepresented with a disparity of 0.10X, while the Hispanic group is underrepresented with a disparity of 0.29X and the Asian group is underrepresented with a disparity of 0.20X. For Language group, the Non-English group is underrepresented with a disparity of 0.50X. In the subsequent analysis, we focus more on the underrepresented groups.

*Impact on False Positive Rate Parity (FPRP):* FPRP is a fairness criteria that ensures that every group has the same FPR. The results in Table III show that FPRP failed for some groups in the dataset. For example, in the Gender attribute, the Female group has a 0.67X disparity, meaning they have a higher FPR than the reference group (Male). In the Ethnicity attribute, the Black group has a 0.76X disparity, meaning they have a higher FPR than the reference group (White). Finally, in the Language attribute, the non-English group has a 0.48X disparity, meaning they have a higher false positive error rate than the reference group (English speaker). This result shows that certain groups have higher false positive error rate than the others and are subject to more unfavourable outcomes.

*Impact on False Negative Rate Parity (FNRP):* FNRP refers to the fairness criteria that ensures that all groups have an equal FNR. The results in Table III shows which groups failed the FNRP test for various attributes. For example, the "Female" gender group has a 1.70X disparity, indicating that they are almost twice as likely to be mistakenly labeled as positive. In the case of ethnicity groups, "Other" has a 0.40X disparity, meaning that they are 40% less likely to be correctly identified as negative compared to the reference group, White. Finally, the "Non-English" language group has a 0.74X disparity, indicating that they are 26% less likely to be correctly identified as negative compared to the reference group, English speakers.

VI. CONCLUSION

This study proposes a ML pipeline for clinical data classification that includes a fairness audit on subgroups based on sensitive attributes. The experiments are performed on MIMIC III dataset and the classification results show the disparities in the outcomes, highlighting the need for better fairness and bias mitigation strategies in ML model building. A limitation of the study is that the experiments are conducted on a subset of data, which may lead to results that are not entirely representative of the overall population or the complete dataset. To achieve more robust and fair AI systems, transparency and accountability can be improved through better documentation, model interpretability, and auditability. Addressing bias and fairness in AI systems is an ongoing challenge that requires a collaborative effort from researchers, policymakers, and practitioners.